\documentclass{article}

\usepackage{caption} 
\usepackage[final,nonatbib]{tackling_climate_workshop_style}
\usepackage[utf8]{inputenc} 
\usepackage[T1]{fontenc}    
\usepackage{hyperref}       
\usepackage{url}            
\usepackage{booktabs}       
\usepackage{amsfonts}       
\usepackage{nicefrac}       
\usepackage{microtype}      

\usepackage{graphicx}
\usepackage{etoolbox}
\usepackage{array,multirow}

\title{Multimodal Wildland Fire Smoke Detection}

\usepackage[T1]{fontenc}
\usepackage[utf8]{inputenc}
\usepackage{babel}
\usepackage{authblk}
\usepackage{blindtext}

\begin{document}

\author{\textbf{Siddhant Baldota}}
\author{\textbf{Shreyas Anantha Ramaprasad}}
\author{\textbf{Jaspreet Kaur Bhamra}}
\author{\textbf{Shane Luna}}
\author{\textbf{Ravi Ramachandra}}
\author{\textbf{Eugene Zen}}
\author{\textbf{Harrison Kim}}
\author{\textbf{Daniel Crawl}}
\author{\textbf{Ismael Perez}}
\author{\textbf{Ilkay Altintas}}
\author{\textbf{Garrison W. Cottrell}}
\author{\textbf{Mai H.Nguyen}\thanks{Corresponding Author: mhnguyen@ucsd.edu}}
\affil{University of California, San Diego} 

\maketitle
\begin{abstract}
Research has shown that climate change creates warmer temperatures and drier conditions, leading to longer wildfire seasons and increased wildfire risks in the United States.  These factors have in turn led to increases in the frequency, extent, and severity of wildfires in recent years.
Given the danger posed by wildland fires to people, property, wildlife, and the environment, there is an urgency to provide tools for effective wildfire management.  Early detection of wildfires is essential to minimizing potentially catastrophic destruction. 
In this paper, we present our work on integrating multiple data sources in SmokeyNet, a deep learning model using spatio-temporal information to detect smoke from wildland fires. Camera image data is integrated with weather sensor measurements and processed by SmokeyNet to create a multimodal wildland fire smoke detection system.
We present our results comparing performance  in terms of both accuracy and time-to-detection for multimodal data vs. a single data source. 
With a time-to-detection of only a few minutes, SmokeyNet can serve as an automated early notification system, providing a useful tool in the fight against destructive wildfires.
\end{abstract}

\section{Introduction}
Research has shown that climate change creates warmer temperatures and drier conditions, leading to longer wildfire seasons and increased wildfire risks in many areas in the United States~\cite{reidmiller2017impacts}\cite{westerling2016increasing}. 
These factors have in turn led to increases in the frequency, extent, and severity of wildfires~\cite{epa}]\cite{wuebbles2017climate}
According to the National Centers for Environmental Information (NCEI)~\cite{noaa-billion}, which keeps track of weather and climate events with significant economic impacts, there have been 20 wildfire events exceeding \$1 billion in damages in the United States from 1980 to 2022, and 16 of those have occurred since 2000~\cite{epa}.
In the western United States, climate change has doubled the forest fire area from 1984 to 2015~\cite{abatzoglou2016impact}.

Given the danger posed by wildland fires to people, property, wildlife, and the environment, there is an urgency to provide tools for effective wildfire management.  Early detection of wildfires is essential to minimizing potentially catastrophic destruction.  Currently, wildfires are detected by humans: scouts trained to be on the lookout for wildfires, residents in an area, or passersby.  We propose here a system for automated wildfire smoke detection to provide early notification of wildfires.

In previous work~\cite{smokeynet2021}, we introduced FIgLib (Fire Ignition Library), a dataset of labeled wildfire smoke images from fixed-view cameras, and SmokeyNet, a novel deep learning architecture using spatio-temporal information from camera imagery to detect smoke from wildland fires.  Here, we extend that work by investigating the use of multiple data sources 
to improve performance in terms of both accuracy and time-to-detection.   Specifically, we integrate weather sensor measurements with camera imagery to create a multimodal wildland fire smoke detection system.

\section{Data}
\label{sec:data}
The work presented in this paper makes use of two different types of data: camera imagery and weather sensor measurements.  
\subsection{FIgLib Data}
Camera imagery comes from the Fire Ignition images Library (FIgLib) dataset~\cite{figlib}. The FIgLib dataset consists of sequences of wildfire images captured from fixed cameras at HPWREN sites~\cite{hpwren_weather}. Each fire contains 80 minutes of MP4 high resolution video feed, with 40 minutes prior to the beginning of the fire and 40 minutes after the beginning of the fire, providing negative and positive samples of fire images, respectively.
After removing out-of-distribution sequences (e.g., night fires), sequences with missing and/or mislabeled annotations, and sequences without matching weather data, a subset of 255 fires remained. This subset was partitioned into 
131 fires for training (51.4\%), 63 fires for validation (24.7\%), and 61 fires for testing (23.9\%). 


\subsection{Weather Data}
\label{subsec:weather_data}
Weather data in the HPWREN\cite{hpwren}\cite{hpwren_weather}, SDG\&E\cite{sdge_weather}, SC-Edison\cite{sc_edison} networks is fetched from weather stations using the Synoptic's Mesonet API\cite{synoptic}. Synoptic is a service that helps in storing and serving weather data. The weather data has 23 attributes, out of which we selected the ones that have under 5\% missing data for the FIgLib time frame of 03 June 2016 to 12 July 2021. These filtered attributes are: air temperature, relative humidity, wind speed, wind gust, wind direction, and dew point temperature. Wind speed and wind direction can be expressed as the radius and angle in a polar coordinate system~\cite{polar}, which are then used to obtain cartesian co-ordinates~\cite{cartesian} $\overrightarrow{u}$ and $\overrightarrow{v}$. This is done in aggregating the weather data.

\section{Methods}
\label{sec:methods}
This section presents methods used for data preparation and model building.

\subsection{Data Preparation}
\label{subsec:data_prep}
The FIgLib data and the weather data have to be pre-processed before being fed into the model. This subsection discusses the procedures that we followed. 

\paragraph{FIgLib Data}
The images of the FIgLIb dataset that we use are resized and cropped to speed up training and to remove clouds (which may cause false positives) into images of size 1040 × 1856 pixels, as this can be then split into $224\times224$ sized tiles. These images are then subjected to random data augmentations and also normalized. Refer to~\cite{smokeynet2021} for details on how these images are processed. 


\paragraph{Weather Data}
For each image in the chosen FIgLib data, we need to get the corresponding weather at the scene captured by the camera. This is done by fetching weather data from the three closest weather stations in the direction that the camera is facing, followed by normalization and aggregation. Also, there is a weather data point available every ten minutes, whereas the images are spaced one minute apart. So we employ linear interpolation to resolve the difference in temporal resolution. 

\subsection{Models}
\label{sub:models}
\begin{figure}
  \centering
  \includegraphics[width=\linewidth]{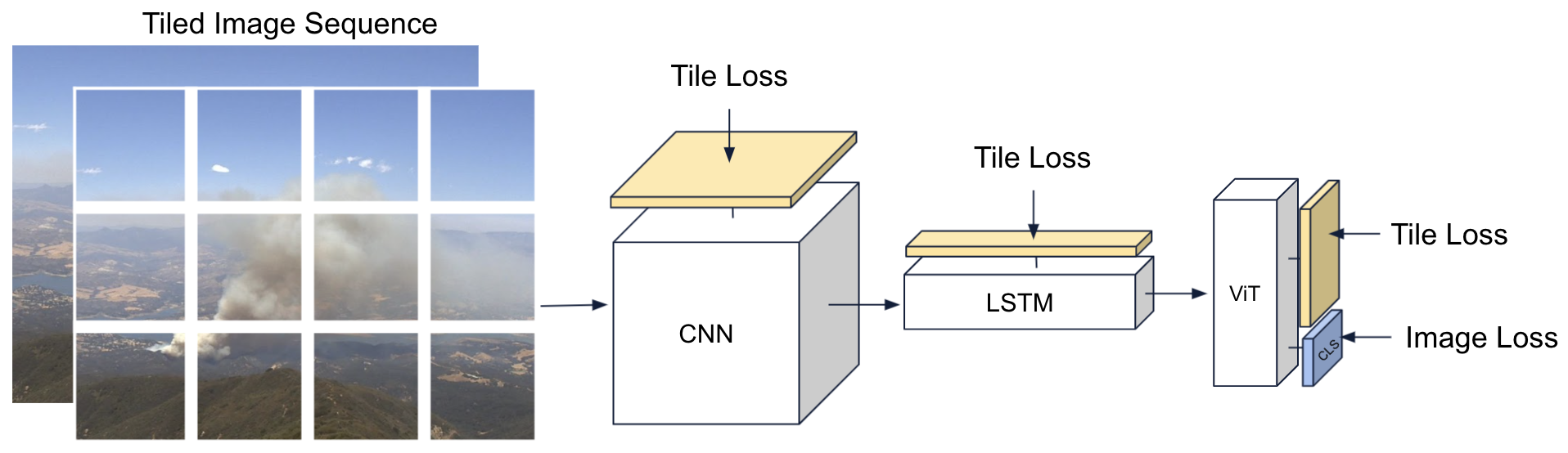}
  \caption{The SmokeyNet architecture takes two frames of the tiled image sequence as input and combines a CNN, LSTM, and ViT. The yellow blocks denote "tile heads" used for intermediate supervision while the blue block denotes the "image head" used for the final image prediction. }
  \label{fig:smokeynetfig}
  \vspace{0.3cm}
\end{figure}

\begin{figure}
  \centering
  \includegraphics[width=\linewidth]{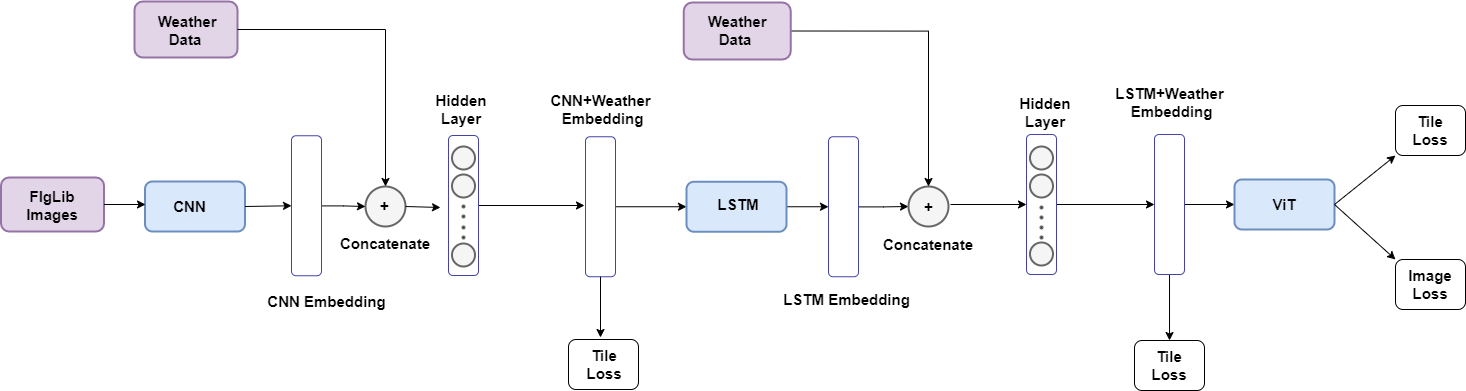}
  \caption{The Multimodal SmokeyNet architecture integrates weather sensor measurements with camera images to perform wildfire smoke detection.}
  \label{fig:linb}
\end{figure}
\paragraph{SmokeyNet}
The baseline model for multimodal wildland fire smoke detection, SmokeyNet, depicted in {Figure 1}, is a spatiotemporal gridded model consisting of three different networks: a convolutional neural network (CNN)~\cite{krizhevsky2012imagenet}, a long short-term memory model (LSTM)~\cite{hochreiter1997long}, and a vision transformer (ViT)~\cite{dosovitskiy2020image}. The input to SmokeyNet is a tiled wildfire image and its previous frame from a wildfire image sequence to account for temporal context. A pre-trained CNN, namely ResNet34~\cite{he2016deep}, extracts feature embeddings from each raw image tile for the two frames independently. These embeddings are passed through an LSTM which assigns temporal context to each tile by combining the temporal information from the current and previous frame. These temporally combined tiles are passed through a ViT, which encodes spatial context over the tiles to generate the image prediction. The outputs of the ViT are spatiotemporal tile embeddings, and a classification (CLS) token that encapsulates the complete image information.\cite{dosovitskiy2020image}. This token is passed through a sequence of linear layers and a sigmoid activation to generate a single image prediction for the current image.
Tile loss is computed using binary cross entropy for the CNN, LSTM, and ViT separately.  Additionally, image loss is computed for the ViT.


\paragraph{Multimodal SmokeyNet}
Figure~\ref{fig:linb} shows the high level architecture for the multimodal SmokeyNet model. For each FIgLib image passed through the model, the corresponding weather vector is also added to the model at two places:  The weather vector is concatenated to the embedding from the CNN and to the embedding from the LSTM. The resulting vector is then passed through a hidden layer. The output from this hidden layer is propagated forward to the next component of SmokeyNet. 
The concatenation of the weather vector to the CNN embedding  increases the dimensions of the weights from the CNN output to the LSTM.  These extra connections are initialized with random weights.  This is similarly done with the weights from the LSTM output to the ViT.
Finally, when the output of the hidden layer is sent through the vision transformer, the final model outputs are tile and image probabilities.

The training procedure for the model happens in two stages. First, the vanilla SmokeyNet model, as described in \cite{smokeynet2021}, is trained for 25 epochs with only camera image data. Using transfer learning, the multimodal model (Figure~\ref{fig:linb}) is initialized with the best weights based on validation loss from the trained vanilla SmokeyNet model.  The model is then further trained with integrated camera image and weather data, as described in further detail in Section~\ref{sec:exps}. 
Training is performed using aggregated CNN tile loss, LSTM tile loss, and ViT tile loss and image loss, as shown in \ref{fig:smokeynetfig}, similar to the vanilla model as described in~\cite{smokeynet2021}.

\section{Experiments}
\label{sec:exps}
As mentioned in Section \ref{sec:data}, we use a train / validation / test split of 131 / 63 / 61 fires (or 10,302 / 4,894 / 4,799 images) for all our experiments. The six weather attributes along with the $\overrightarrow{u}$ and $\overrightarrow{v}$ components as described in Section~\ref{subsec:weather_data} constitute the vector, resulting in a weather vector of length eight.  First, we run experiments on the vanilla SmokeyNet model to establish a baseline, and then train the multimodal SmokeyNet model with the integrated weather data. 

\paragraph{SmokeyNet Baseline}
As described previously, we need a baseline with which to compare the multimodal SmokeyNet model. 
For the baseline model, we take the original trained SmokeyNet model and train it for an additional 25 epochs.

\paragraph{Adding Weather Data}
Using the transfer learning approach and integrated camera image and weather data as described in Section ~\ref{sub:models} 
we train the multimodal SmokeyNet model shown in Figure~\ref{fig:linb} for 25 epochs with early stopping. 
Additionally, to verify that the addition of the weather data is adding some useful information to the model, we also run experiments by passing random weather tensors of the same size, drawn from a normal distribution.

For all experiments, we use the best values for the hyperparameters as mentioned in \cite{smokeynet2021}, i.e, a learning rate of 1e-3, weight decay of 1e-3, image resizing of 90\%, no dropout, image binary cross entropy loss with positive weight of 5, and a batch size of 2.
To give more weighting to the weather data, we use a replication factor of ten, which means that the weather vector of size eight is replicated $8\times10$ and then concatenated to the CNN/LSTM embedding. 

\section{Results \& Discussion}
\label{sec:disc}
Table \ref{tab:exp_results} provides a summary of the experiments.  For each experiment, we report the accuracy, precision, recall, F1 score, and time-to-detection (TTD).  For each row in the table, the reported scores are the average and standard deviation over eight runs.


\begin{table}[h!]
\resizebox{\columnwidth}{!}{%
\begin{tabular}{|l|ll|cc|cc|cc|cc|}
\hline
\multicolumn{1}{|c|}{\multirow{2}{*}{Model}} &
  \multicolumn{2}{c|}{TTD (minutes)} &
  \multicolumn{2}{c|}{Accuracy} &
  \multicolumn{2}{c|}{F1} &
  \multicolumn{2}{c|}{Precision} &
  \multicolumn{2}{c|}{Recall} \\ \cline{2-11} 
\multicolumn{1}{|c|}{} &
  \multicolumn{1}{c|}{Mean} &
  \multicolumn{1}{c|}{SD} &
  \multicolumn{1}{c|}{Mean} &
  SD &
  \multicolumn{1}{c|}{Mean} &
  SD &
  \multicolumn{1}{c|}{Mean} &
  SD &
  \multicolumn{1}{c|}{Mean} &
  SD \\ \hline
SmokeyNet &
  \multicolumn{1}{l|}{4.70} &
  0.90 &
  \multicolumn{1}{c|}{\textbf{80.12}} &
  1.47 &
  \multicolumn{1}{c|}{77.52} &
  2.39 &
  \multicolumn{1}{c|}{\textbf{90.43}} &
  1.66 &
  \multicolumn{1}{c|}{68.00} &
  4.42 \\ \hline
\begin{tabular}[c]{@{}l@{}}SmokeyNet with \\ Random Weather\end{tabular} &
  \multicolumn{1}{l|}{4.88} &
  0.96 &
  \multicolumn{1}{c|}{79.50} &
  0.77 &
  \multicolumn{1}{c|}{76.90} &
  1.31 &
  \multicolumn{1}{c|}{89.40} &
  1.51 &
  \multicolumn{1}{c|}{67.53} &
  2.63 \\ \hline
\begin{tabular}[c]{@{}l@{}}SmokeyNet with \\ Weather\end{tabular} &
  \multicolumn{1}{l|}{\textbf{3.66}} &
  0.77 &
  \multicolumn{1}{c|}{79.97} &
  1.18 &
  \multicolumn{1}{c|}{\textbf{78.18}} &
  1.68 &
  \multicolumn{1}{c|}{87.07} &
  2.16 &
  \multicolumn{1}{c|}{\textbf{71.07}} &
  3.54 \\ \hline
\end{tabular}%
}
\caption{Mean and standard deviation (SD) of Time-to-Detection (TTD), Accuracy, F1, Precision, and Recall metrics on the test set over eight runs. SmokeyNet is the baseline model without weather data.  SmokeyNet with Random Weather uses the multimodal SmokeyNet architecture but with random numbers for the weather vector.  SmokeyNet with Weather is multimodal SmokeyNet with actual weather data.}
\label{tab:exp_results}
\end{table}

From these results, we observe that both F1 and time-to-detection improved with the multimodal version of SmokeyNet over the baseline version with just camera imagery.  Averaged over eight runs, F1 improved slightly. And the time-to-detection metric improved by approximately 1 minute, or 22\%.  
Additionally, the standard deviation for both F1 and time-to-detection also decreased.
Thus, the multimodal SmokeyNet model not only improves the time-to-detection and F1 on average, but also offers more stability in these metrics across fire sequences.

Our results demonstrate that SmokeyNet can effectively process multiple data sources for wildland fire smoke detection, boosting detection performance in terms of both F1 and time-to-detection over the baseline with a single data source.  
With a time-to-detection within a few minutes, SmokeyNet can be used as an automated early notification system, providing a useful tool in the fight against destructive wildfires.

For future work, we will analyze our results to gain insights into scenarios in which adding weather data improves performance.  We will also investigate approaches to make use of unlabeled data to further improve detection performance.  Additionally, we will explore methods to optimize the model's compute and memory resource requirements.  Ultimately, our goal is to embed SmokeyNet into edge devices to enable insight at the point of decision for effective real-time wildland fire smoke detection.

\clearpage
\section{Acknowledgements}
We would like to thank the Meteorology Team at SDG\&E for their valuable feedback and support.
This work was supported in part by funding from SDG\&E, and NSF award numbers 1730158, 2100237, 2120019 for Cognitive Hardware and Software Ecosystem Community Infrastructure (CHASE-CI).

\bibliographystyle{unsrt}
\bibliography{sources}
\nocite{*}

\clearpage

\end{document}